%% file: main.tex
\documentclass[10pt,twocolumn,letterpaper]{article}

\usepackage{cvpr}              

\usepackage{times}
\usepackage{epsfig}
\usepackage{graphicx}
\usepackage{amsmath}
\usepackage{amssymb}
\usepackage{booktabs}
\usepackage{float}
\usepackage{caption}
\usepackage{pifont}
\usepackage{stfloats}
\usepackage{multirow}
\usepackage{tabularx}
\usepackage{anyfontsize}
\usepackage[accsupp]{axessibility}
\input{preamble}

\definecolor{cvprblue}{rgb}{0.21,0.49,0.74}
\usepackage[pagebackref,breaklinks,colorlinks,citecolor=cvprblue]{hyperref}

\def\paperID{41}
\def\confName{EarthVision}
\def\confYear{2024}

\title{Sat2Cap: Mapping Fine-Grained Textual Descriptions from Satellite Images}

\author{Aayush Dhakal\textsuperscript{1}, Adeel Ahmad\textsuperscript{1,2}, Subash Khanal\textsuperscript{1}, Srikumar Sastry\textsuperscript{1}, Hannah Kerner\textsuperscript{3}, Nathan Jacobs\textsuperscript{1}\\
{\normalsize \textsuperscript{1}Washington University in St. Louis, \textsuperscript{2}Taylor Geospatial Institute, \textsuperscript{3}Arizona State University} \\
{\{\tt\small a.dhakal, aadeel, k.subash, s.sastry, jacobsn\}@wustl.edu, hkerner@asu.edu}, 
}

\begin{document}
\maketitle
\input{sec/0_abstract}    
\input{sec/1_intro}
\input{sec/2_related}

\input{sec/3_method}

\input{sec/4_experiments}
\input{sec/5_conclusion}
\input{sec/X_suppl}

\clearpage
{
    \small
    \bibliographystyle{ieeenat_fullname}
    \bibliography{main}
}

\end{document}

%% file: preamble.tex
\usepackage[dvipsnames]{xcolor}


%% file: sec/0_abstract.tex
\begin{abstract}
We propose a weakly supervised approach for creating maps using free-form textual descriptions. We refer to this work of creating textual maps as zero-shot mapping. Prior works have approached mapping tasks by developing models that predict a fixed set of attributes using overhead imagery. However, these models are very restrictive as they can only solve highly specific tasks for which they were trained. Mapping text, on the other hand, allows us to solve a large variety of mapping problems with minimal restrictions. To achieve this, we train a contrastive learning framework called Sat2Cap on a new large-scale dataset with 6.1M pairs of overhead and ground-level images. For a given location and overhead image, our model predicts the expected CLIP embeddings of the ground-level scenery. The predicted CLIP embeddings are then used to learn about the textual space associated with that location. Sat2Cap is also conditioned on date-time information, allowing it to model temporally varying concepts over a location. Our experimental results demonstrate that our models successfully capture ground-level concepts and allow large-scale mapping of fine-grained textual queries. Our approach does not require any text-labeled data, making the training easily scalable. The code, dataset, and models will be made publicly available.
\end{abstract}

%% file: sec/1_intro.tex
\begin{figure*}[t]
\begin{center}
\includegraphics[width=0.95\linewidth]{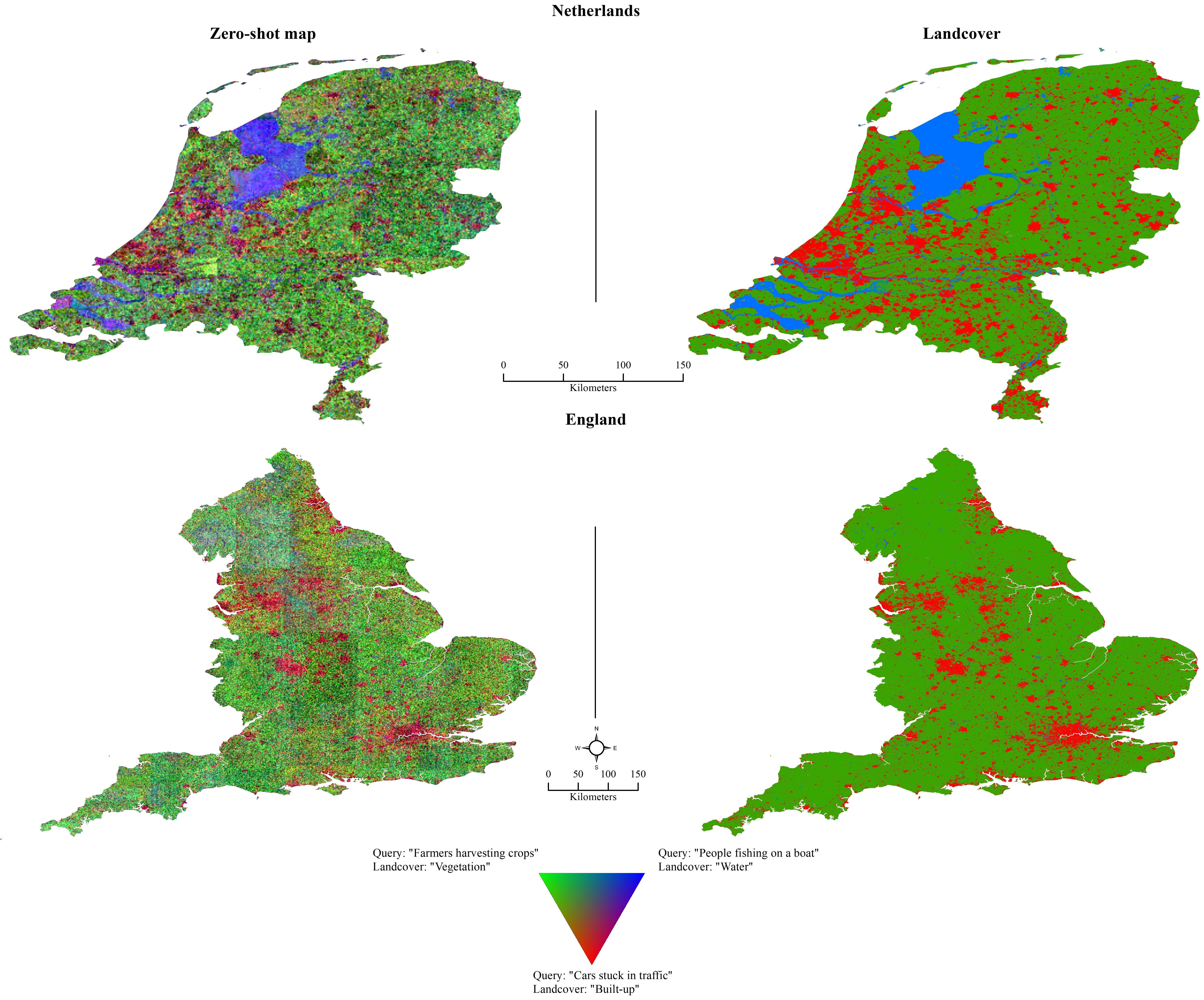}
\end{center}
   \caption{\textbf{Country-level maps of textual descriptions}: (Col 1-2) shows the country-level maps created using Sat2Cap for three prompts: ``Cars stuck in traffic", ``People fishing on a boat," and ``Farmers harvesting crops." We compare the predicted zero-shot maps with landcover maps of the region.}  
\label{fig:generic_maps}
\end{figure*}

\section{Introduction}
\label{sec:intro}

Maps are a fundamental data product for a wide variety of domains. Traditional map-making involved extensive ground-based surveys. However, such methods are extremely time-consuming, expensive, and labor-intensive. As a result, overhead remote-sensing imagery has emerged as an important data modality for map creation. Machine learning methods have enabled scalable, accurate mapping of attributes using overhead imagery. The current paradigm of prior methods is to learn models that leverage the visual cues from overhead images to predict specific pre-defined attributes (e.g., a fixed set of land cover classes). Persello et al~\cite{persello2022deep} describe the numerous applications of deep learning in addressing the Sustainable Development Goals (SDGs), including crop monitoring, deforestation mapping, wildfire monitoring, and more. Salem et al.~\cite{salem2020learning} used overhead images to map transient attributes~\cite{laffont2014transient} and scene categories~\cite{zhou2017places} across large regions, while Streltsov et al.~\cite{streltsov2020estimating} predicted residential building energy consumption using overhead imagery. Similarly, Bency et al.~\cite{bency2017beyond} also used satellite images to map housing prices. 

All these prior methods focused on learning some specific pre-defined attributes. These attribute-specific models are highly restrictive as they cannot map anything beyond their preset list of variables. To overcome this limitation, we introduce a framework that enables the mapping of fine-grained textual descriptions of concepts that are observable only on the ground. Our approach allows the mapping of any concept that can be expressed in natural language and, thus, serves as a general framework for zero-shot mapping. For example: using our model, one can create a map of concepts like ``harvesting crops" or ``busy streets" without training any task-specific models.

Modeling the relationship between text and images is a well-studied problem in deep learning. Numerous  methods~\cite{radford2021learning, li2021align, yao2021filip} have been proposed to learn the relationship between these two modalities. Models such as CLIP~\cite{radford2021learning} and ALBEF~\cite{li2021align} are trained on a large database of captioned images to learn a multimodal embedding space that unifies vision and text. However, we observe that directly using these models on overhead imagery leads to the collapse of representations to a few coarse textual concepts like city, beach, island, etc. Overhead images capture a broad perspective of a geolocation but offer limited insight into the intricate concepts and dynamics within the location. Hence, directly using overhead images with such models for text-driven mapping would only allow us to map coarse-level textual concepts. On the other hand, ground-level images and their respective CLIP embeddings provide detailed and fine-grained concepts of a location. Yet, several challenges hinder the direct utilization of ground-level imagery for mapping tasks. Firstly, ground-level images are sparsely available, i.e., obtaining a ground-level image for every location on Earth is not feasible. Secondly, the coverage and quality of a ground-level image for the same location can have large variations, which could introduce unwanted variations during inference.

To address these issues, we present a weakly-supervised cross-view approach for learning fine-grained textual concepts for geographic locations. In our work, ``fine-grained" refers to concepts that are observable in ground-level images but are hard to infer from the low-resolution view of satellite images. To do this, we first create a large-scale dataset with paired overhead and ground-level images. Our dataset uses a subset of the YFCC100M~\cite{thomee2016yfcc100m}. More details about the dataset are presented in Section~\ref{dataset}. Using this paired dataset, we learn the CLIP embeddings of the ground-level images at a given location. CLIP embeddings of ground-level images can describe detailed textual concepts of that location. Using the overhead image, our Sat2Cap model learns to predict the expected CLIP embeddings of the ground-level scene. Compared to the CLIP embeddings, Sat2Cap embeddings tend to capture more fine-grained textual concepts for a given geolocation.

To account for the temporal associations between various concepts and locations, our model is conditioned on temporal data, specifically, the date and time stamps from the Flickr imagery. This allows our model to learn concepts that can be dynamically adapted to different date and time settings. Our method is also weakly-supervised and thus does not require any text labels. To summarize, these are the primary contributions of our work:
\begin{itemize}
  \item A weakly-supervised approach for learning fine-grained textual concepts of geographic locations
  \item A zero-shot approach for creating large-scale maps from textual queries as seen in Figure~\ref{fig:generic_maps}
  \item A new large-scale cross-view dataset 
  \end{itemize}

%% file: sec/2_related.tex
\section{Related Works}
\label{sec:related}
\subsection{Deep Learning Based Mapping}
The ability to map attributes of interest is a fundamental task in Remote Sensing and has wide-ranging implications for achieving SDGs. Deep Learning methods have been used extensively~\cite{persello2022deep,kavvada2020towards,behrens2018multi,zong2019deepdpm,onishi2021explainable,greenwell2018goes} in recent years to make mapping efficient and scalable. Alhassan et al.~\cite{alhassan2020deep} finetuned imagenet pretrained models to make landcover predictions. Similarly~\cite{9553499,feizizadeh2023machine} leveraged large-scale annotated data from different sensors to improve land use and landcover classification using deep learning methods. Other works specifically focus on mapping visual attributes. For instance,~\cite{workman2017understanding} used features from both overhead and ground-level imagery and introduced a cross-view approach to map scenic-ness. Later works focused on creating dynamic maps: ~\cite{workman2020dynamic,salem2020learning} conditioned their model on temporal information along with overhead images to learn dynamic concepts for a given location. However, across this huge research area, the prevailing paradigm is to create task-specific models over a fixed set of attributes. We attempt to generalize the mapping process to \textit{any} attribute by introducing a framework to create maps of free-form textual prompts. 

\subsection{Vision-Language Pretraining}
Vision-Language (VL) models have shown great promise in their ability to model complex relationships between the vision and text space. ConVIRT~\cite{zhang2022contrastive} and VirTex~\cite{desai2021virtex} both introduced methods that used image-text pairs to learn rich visual representations. CLIP~\cite{radford2021learning} demonstrated the results of VL pretraining on a large-scale dataset (400M pairs) and validated the efficacy of large-scale VL pretraining for several downstream tasks. Florence~\cite{yuan2021florence} and ALIGN~\cite{jia2021scaling} further increased the scale of data by training on 900M and 1.8B pairs, respectively. Other works~\cite{li2021align,yao2021filip,yang2022unified,yang2022vision} have since focused on learning better VL embedding space. With the existence of these powerful pretrained VL models, many researchers have utilized their embedding spaces to solve specific downstream tasks. CLIPCap~\cite{mokady2021clipcap} and~\cite{cho2022fine} used CLIP space to generate image captions. Models like~\cite{ramesh2022hierarchical,nichol2021glide,wang2022clip} utilized the CLIP space for text-to-image generation. 

Recently, there has been work in creating image-text datasets of overhead images and captions for VL pertaining of geospatial models. ChatEarthNet~\cite{yuan2024chatearthnet} created a dataset with paired image and text captions using ChatGPT. Similarly, SkyScript~\cite{wang2023skyscript} used OpenStreetMap (OSM) data to create overhead image and text-paired datasets for VL pretraining. However, both approaches only gather coarse textual information for a given location using either the low-resolution visual cues from an overhead image or fixed preset tags from OSM data.

Our approach fundamentally differs from prior work as we try to capture the more intricate concepts occurring at the ground-level of a given location by leveraging the corresponding crowdsourced images. Utilizing the ground-level images uploaded from a location allows us to access extremely fine-grained information of that location that is not distillable solely from satellite images or OSM data.

%% file: sec/3_method.tex
\begin{figure*}[!t]
\begin{center}
\includegraphics[width=\linewidth, scale=0.3]{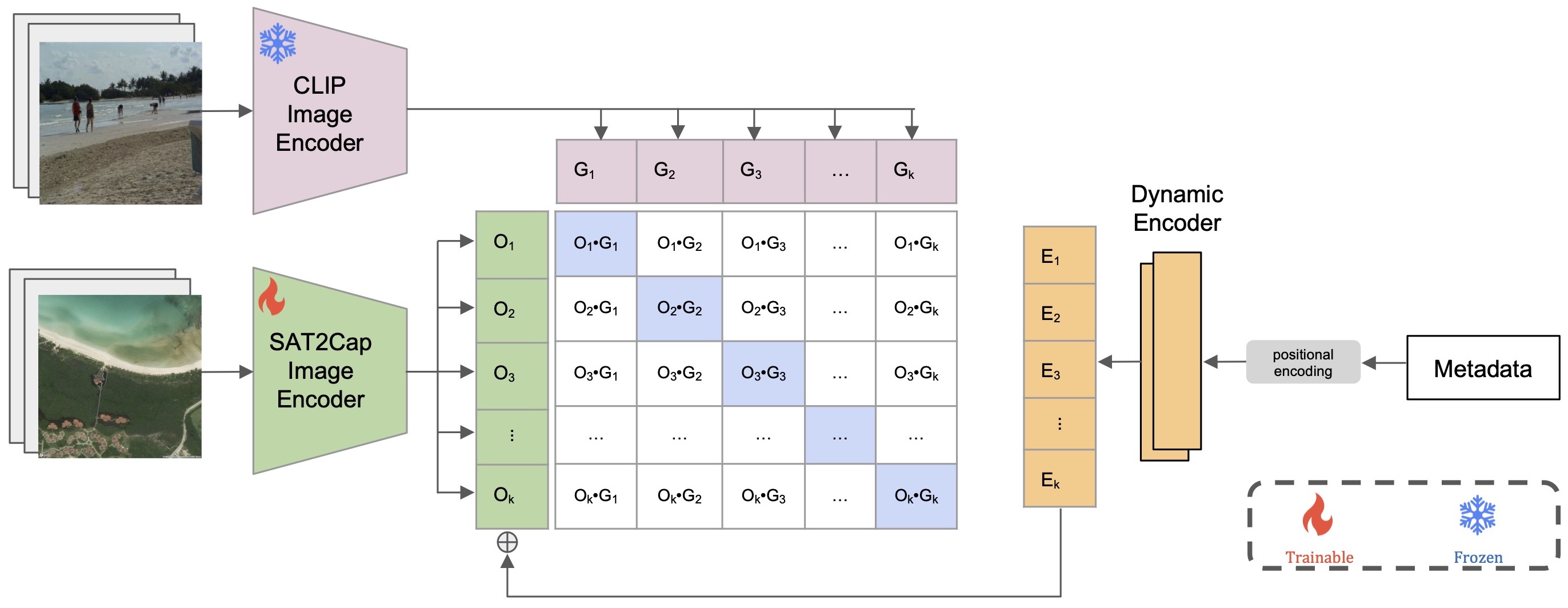}
\end{center}
   \caption{\textbf{Sat2Cap Framework:} The frozen CLIP Image Encoder takes as input the ground-level images and generates their CLIP embeddings. The trainable Sat2Cap Image Encoder takes as input the overhead images, and the Dynamic Encoder takes as input the date, time, and location information. The respective overhead image embeddings and meta-information embeddings are added element-wise, and the resulting embeddings are contrastively trained with the CLIP embeddings of the ground-level images.}  
\label{fig:Sat2Cap}
\end{figure*}

\section{Method}
Our objective is to learn an embedding space that describes the expected ground-level scene given a geographic location and an overhead image. We have ground-level images $\{g_1, g_2, ...g_n\}$, corresponding overhead images $\{o_1, o_2, ...o_n\}$, and respective metadata for the ground-level images $\{e_1, e_2, ...e_n\}$. Each $e_i$ contains the latitude and longitude information of the sample, as well as the date and time for when the ground-level image was captured. We also have a CLIP image encoder $f_{\theta}$ that generates CLIP embeddings for a given ground-level image.

\subsection{Dataset}
\label{dataset}
We created a large-scale cross-view dataset to train our model. The ground-level images in the dataset are taken from the YFCC100M~\cite{thomee2016yfcc100m} dataset. The YFCC100M dataset contains 99.3 million images, which are collected from Flickr. Our cross-view dataset uses a smaller sample from this collection which excludes all US imagery. We filter out all images that are not geotagged. The resulting dataset contains 6.1M images. Each of these images has a geolocation, timestamp, and other meta-information. For each image, we download an overhead image centered at its location. Specifically, we use the Bing Maps API to download $800$x$800$ patch satellite images at $0.6m/px$ resolution. We randomly sampled 100k images to be used for testing.

\subsection{Approach}
We initialize our Sat2Cap image encoder $g_\theta$ with the weights of $f_\theta$. A batch of ground-level images $\{g_1, g_2, ..., g_k\}$ is passed through the CLIP encoder to get the ground-level CLIP embeddings. These embeddings serve as the target for alignment. A batch of corresponding overhead images $\{o_1, o_2, ..., o_k\}$ is passed through the Sat2Cap image encoder to obtain the respective embeddings:  
\begin{equation}
    G_i = f_\theta{(g_i)}
\end{equation}
\begin{equation}
    O_i = g_\theta{(o_i)}
\end{equation}    

To align the overhead image embeddings with the ground-level CLIP embeddings, we contrastively train our model using the InfoNCE~\cite{oord2018representation} loss as follows:
\begin{equation}
    L = \frac{1}{k}\sum_{i=0}^{k} -\log\frac{exp(O_i \cdot G_i / \tau)}{\sum_{j=0}^{k} exp(O_i \cdot G_j / \tau) }
\end{equation}

We also keep a queue $Q$ and fill it with CLIP embeddings of ground images. Here, $|Q|>>k$ and the embeddings from the queue are used as additional negative samples for our contrastive objective. As in MoCo~\cite{he2020momentum}, the queue is continuously updated during the training with the most recent batch. Minimizing this objective minimizes the distance between co-located overhead and ground-level images in the CLIP space. It is worth noting that throughout the training process, the CLIP image encoder is frozen. Hence, with our training procedure, we allow the overhead images to move close to images from their respective ground-level scene in the CLIP space. As a byproduct, the overhead images also move closer to the textual descriptions of ground-level images, which we are ultimately interested in mapping.

\subsection{Learning Dynamic Concepts of Places}

Many ground-level concepts are temporally dependent. Concepts like ``crowded street", or ``snowy place" can dramatically vary based on the exact time we query about them. To model such dynamic concepts, we condition Sat2Cap on the timestamps of the ground-level images.

For each sample, we extract the year, month, day, and hour in which the ground image was taken. We also add the geolocation information to provide a stronger signal to the model. We encode this meta-information using sin-cos encoding and pass it through a shallow, fully connected network, which we call the Dynamic Encoder $(h_\theta)$. The output from $h_\theta$ is added element-wise to the output from the Sat2Cap encoder before computing the objective.
\begin{equation}
    S_i = O_i + E_i
\end{equation}
where $O_i$ is the output from the Dynamic Encoder. Now the objective function changes to:
\begin{equation}
\label{eq:finaleq}
   L_{dynamic} = \frac{1}{k}\sum_{i=0}^{k} -\log\frac{exp(S_i \cdot G_i / \tau)}{\sum_{j=0}^{k} exp(S_i \cdot G_j / \tau) } 
\end{equation}
where $S_i$ is the sum of outputs from the Image Encoder and Dynamic Encoder



Our complete framework is shown in Figure~\ref{fig:Sat2Cap}. To prevent overfitting to the meta-information, we implement random dropout of the Dynamic Encoder during training. Our experiments from Section~\ref{retrieval_section} show that this addition significantly improves the performance of our model.
\newcommand{\cmark}{\ding{51}}%
\newcommand{\xmark}{\ding{55}}%

\begin{table*}
 \begin{center}
 \small
    \begin{tabular}{c c c c |c c c|c c c}
 \hline
 \multicolumn{4}{c|}{Method} &
 \multicolumn{3}{c|}{Overhead2Ground (10K)} &
 \multicolumn{3}{c}{Ground2Overhead (10K)} \\
 
 \hline
 Model &Meta/Training &Dropout & Meta/Inference & R@5$\uparrow$ & R@10$\uparrow$ & Median-R$\downarrow$ & R@5$\uparrow$ & R@10$\uparrow$ & Median-R$\downarrow$\\
 \hline
 CLIP & - &- & - & 0.007 & 0.013 & 1700 & 0.108 & 0.019 & 2857 \\
 \hline
 \multirow{4}{2em}{ours} & \xmark & \xmark & \xmark & 0.398 & 0.493 & 15 & 0.356 & 0.450 & 11\\
 & \cmark & \xmark & \xmark & 0.322 & 0.413 & 34 & 0.254 & 0.343 & 20\\
 & \cmark& \xmark & \cmark & 0.368 & 0.467 & 23 & 0.298 & 0.398 & 13\\
 & \cmark& \cmark & \xmark & 0.467 & 0.564 & 13.5 & 0.366 & 0.462 & 7\\
 & \cmark& \cmark & \cmark & \textbf{0.493} & \textbf{0.591} & \textbf{12} & \textbf{0.390} & \textbf{0.482} & \textbf{6}\\

\end{tabular}
\end{center}
  \caption{\textbf{Cross-view retrieval performance of Sat2Cap model:} The table shows that CLIP performs poorly for the task of cross-view retrieval. Moreover, we study the performance of our model under various settings. The \textbf{Meta/Training} column ablates the impact of adding meta-information in training. The \textbf{Dropout} column ablates the impact of randomly dropping out the Dynamic Encoder in training. The \textbf{Meta/Inference} column ablates the impact of adding meta-information in inference. Our experiments show that using all three achieves the best performance.}
\label{table:retrieval}
\end{table*}
\begin{figure*}[t]
\begin{center}
\includegraphics[width=0.95\linewidth,scale=0.1]{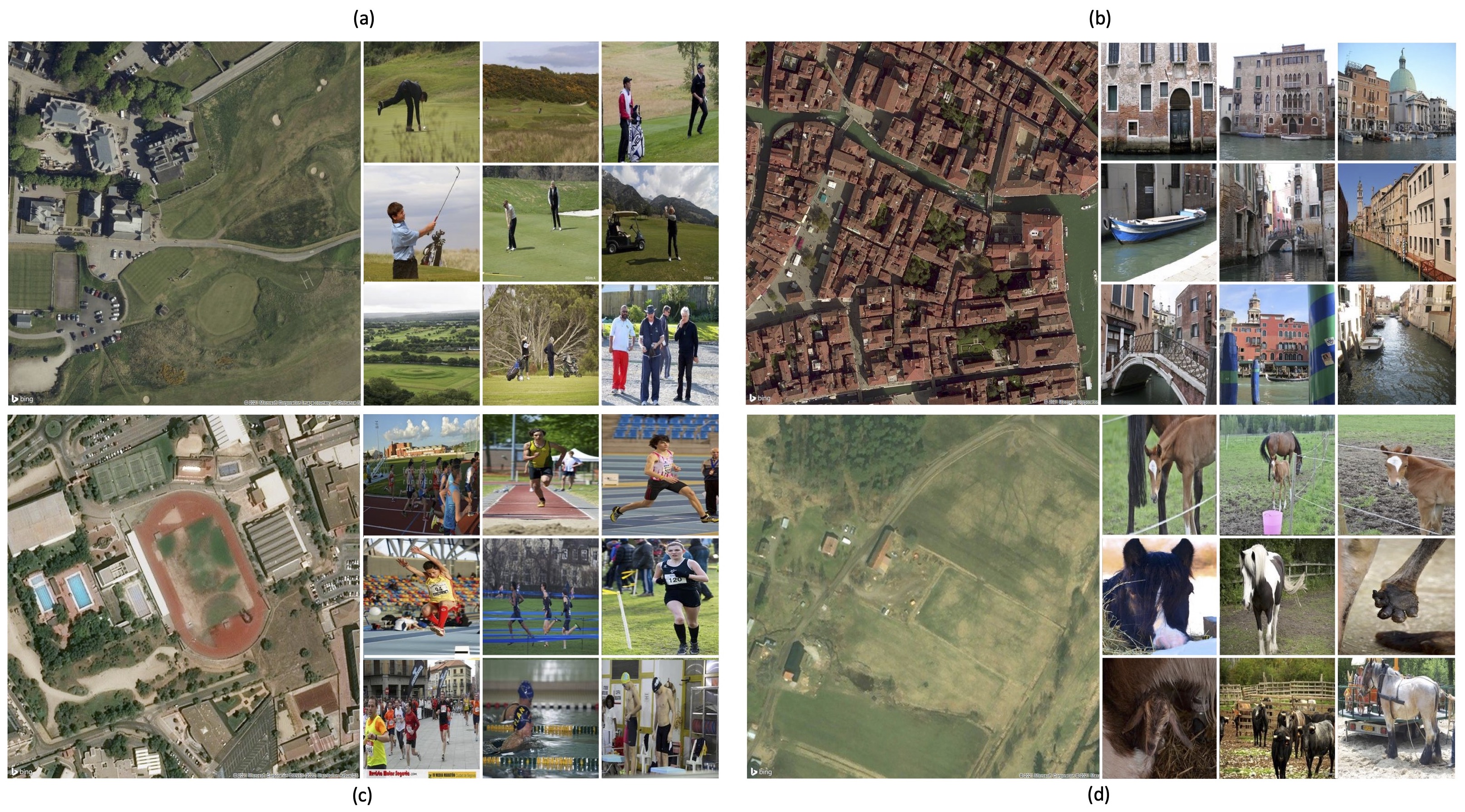}
\end{center}
   \caption{\textbf{Top-9 overhead-to-ground image retrieval:} We use the Sat2Cap embeddings of the overhead images and CLIP embeddings of the ground-level images and show the 9 closest ground-level images retrieved for a query overhead image. The retrieval was performed from a gallery of 10,000 samples.}  
\label{fig:retrieval_results}
\end{figure*}

\begin{figure}[t]
\begin{center}
\includegraphics[width=0.95\linewidth]{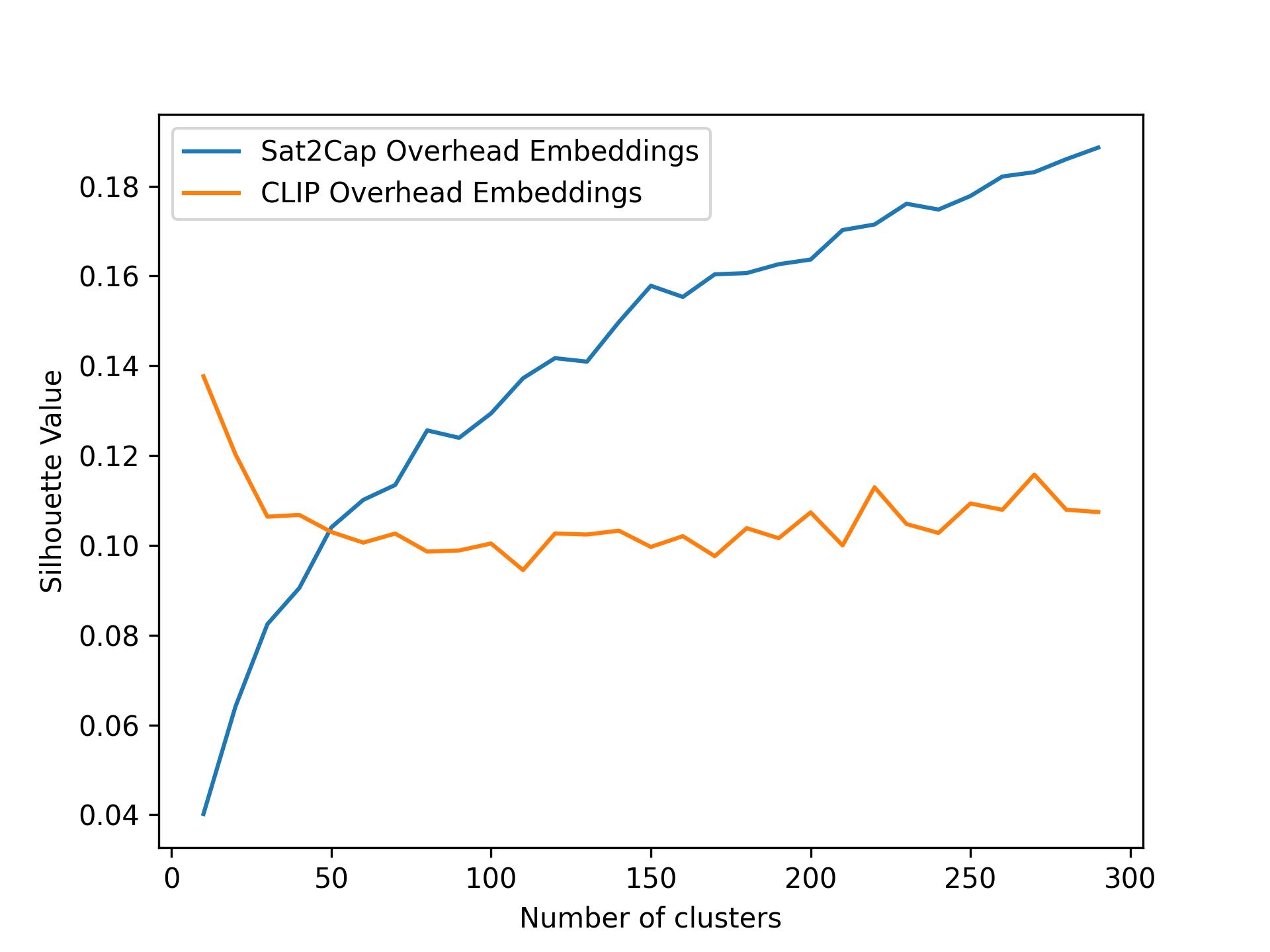}
\end{center}
   \caption{\textbf{Silhouette value of CLIP embedding vs.\ Sat2Cap embedding clusters:} We use k-means clustering with identical parameters to get the clusters for different values of k. Results show that Sat2Cap embeddings can be well separated into a larger number of clusters than the corresponding CLIP embeddings.}  
\label{fig:sill}
\end{figure}

\begin{figure*}[t]
\begin{center}
\includegraphics[width=0.95\linewidth]{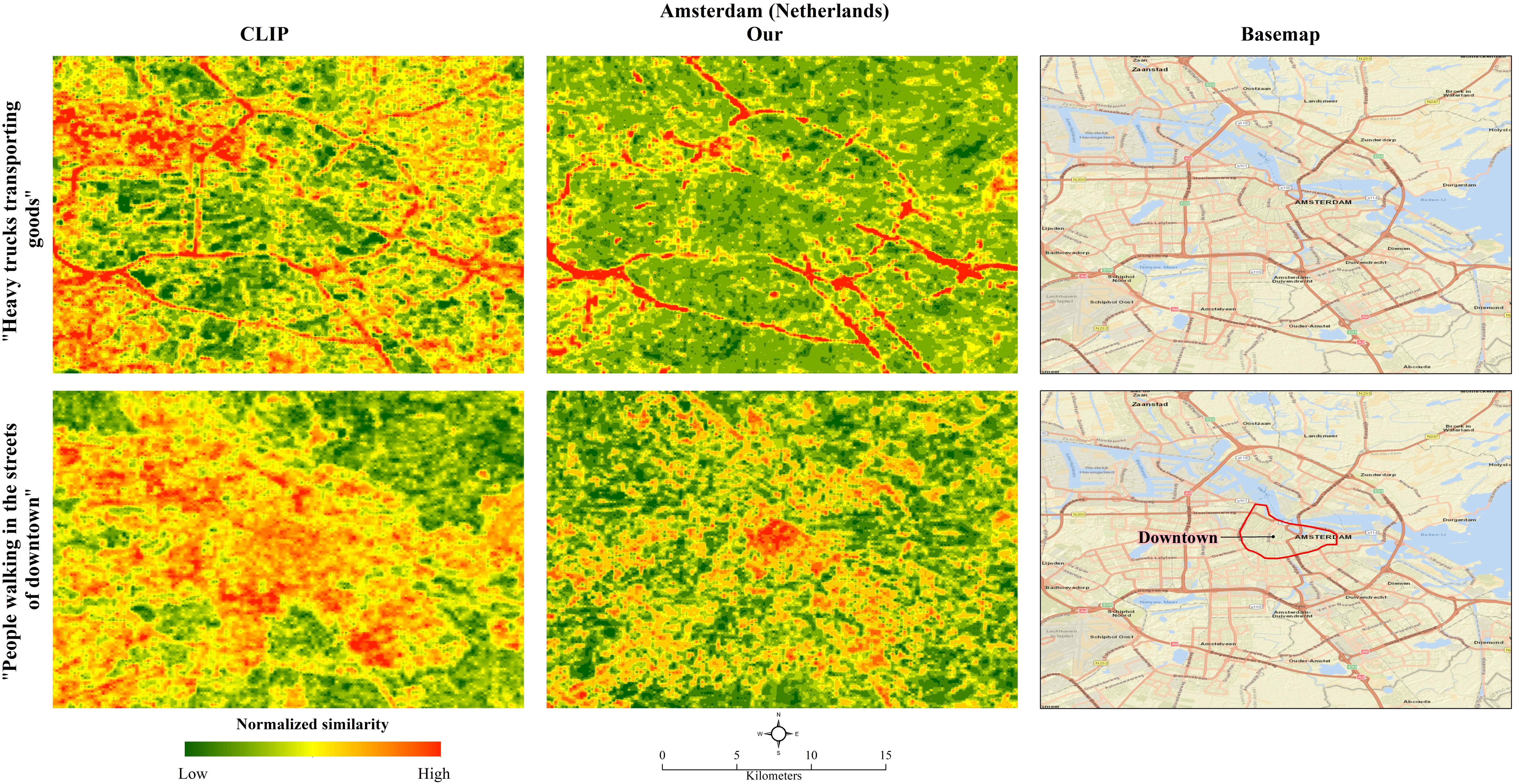}
\end{center}
   \caption{\textbf{Fine-grained maps CLIP vs Sat2Cap:} We create zero-shot maps on a city level using CLIP and Sat2Cap for two prompts: ``Heavy trucks transporting goods" and ``People waking in the streets of downtown." Compared to CLIP, Sat2Cap activations are more localized to the appropriate regions for a given prompt. Sat2Cap is better at distinguishing between fine-grained concepts like \textbf{highway} and \textbf{downtown street}.}   
\label{fig:fg_maps}
\end{figure*}

%% file: sec/4_experiments.tex

\section{Experiments and Results}
Our model learns a powerful geo-text embedding space that can be used for a variety of applications. We describe 4 experiments to demonstrate the efficacy of our model.

\subsection{Cross-View Image Retrieval}
\label{retrieval_section}
In this experiment, we show that our model learns a strong relationship between co-located overhead images and ground images in the CLIP space. We randomly sample 10,000 image pairs from the test set for this experiment. First, we predict the Sat2Cap embeddings for all overhead images and the corresponding CLIP embeddings for all the ground-level images in the test set. We then compute top-k (or R@k) and median rank metrics between the Sat2Cap overhead embeddings and the CLIP ground embeddings. The top-k metric measures how often the ground truth falls within the top-k closest images for a given query image. Table~\ref{table:retrieval} shows all of the cross-view retrieval results.

As a baseline, we use the retrieval results using CLIP overhead embeddings and CLIP ground embeddings. The cross-view retrieval for CLIP model is extremely low with R@10 score of 0.013 and a median rank of 1700. These low scores suggest that the overhead image CLIP embeddings do not contain high-frequency information about the ground-level scene. For our model, we first experiment without using the Dynamic Encoder. Just by contrastively training the Sat2Cap image encoder with ground-level CLIP embeddings, we achieve a high R@10 score of 0.493 and a median rank of 15, as seen in Table~\ref{table:retrieval}. 

All remaining experiments are conducted on models trained using the Dynamic Encoder. Table~\ref{table:retrieval} shows that initially, the retrieval scores drop when using the Dynamic Encoder. This happens because the model starts to overfit on the meta-information, ignoring important cues from the overhead images. We see a further 5.4\% drop in R@10 metrics when we use meta-information in training but remove it during inference. To reduce the possibility of overfitting, we randomly drop the Dynamic Encoder during training. Simply adding dropout during training increases the R@10 score by 12.4\%. Furthermore, removing meta-information during inference is less severe (2.7\%) when using dropout. Hence, our model achieves good cross-view retrieval scores even if meta-information is not provided during inference. 

Figure~\ref{fig:retrieval_results} shows the 9 closest images retrieved from a given overhead image. We see that our model is able to retrieve ground-level images by relating concepts rather than direct visual matching. For example, in (d), our model relates farmland with cattle and livestock, which are concepts that semantically match the location but are not visible in the overhead image.
Sat2Cap is also capable of encoding temporal information in the embeddings whose results are shown in the supplementary material. 

Our results highlight that CLIP space, in itself, is remarkably poor at learning the relationship between a location and the corresponding ground-level scene. The low cross-view retrieval scores imply that we cannot accurately reason about ground-level scene using the CLIP embeddings of co-located overhead images. On the other hand,  the high cross-view retrieval scores of Sat2Cap suggest that the Sat2Cap embedding of a location approximates the CLIP embeddings of the images at the ground-level. This ultimately results in an emergent alignment between location and textual descriptions of the ground level.

\subsection{Embedding Space Comparison: CLIP vs.\ Sat2Cap}
In Section~\ref{retrieval_section}, we used retrieval scores to draw conclusions about the Sat2Cap and CLIP space. Here, we highlight the differences in Sat2Cap and CLIP space more explicitly. In our work, the term ``fine-grained" refers to the concepts that are easily visible at the ground level but are hard to infer from overhead images. For example, an overhead image over a location might tell us that it is a city, but ground-level images of that location capture detailed information such as how crowded the place is, if there are many restaurants around that area, does the location regularly hosts street festivals, and so on. Our method learns a model that takes an overhead view of a location and predicts a representation of ground-level images at that location in CLIP space.

We hypothesize that, for given overhead images, Sat2Cap learns diverse concepts while CLIP collapses to a few coarse concepts. To examine this, we run k-means clustering for both CLIP and Sat2Cap overhead image embeddings. We then compute the silhouette value for the resulting clusters, which tells us the quality of these clusters. This value ranges from -1 to +1, and it tells us how well the embeddings are separated/clustered for a given value of $k$; higher values indicate better clusters. Figure~\ref{fig:sill} shows the plot of silhouette value for different values of $k$. CLIP embeddings are well clustered for small values of $k$ but gradually worsen as we increase the number of clusters. This suggests that CLIP embeddings collapse to only a few concepts and are not further separable to reason about more diverse concepts. On the contrary, Sat2Cap embeddings perform poorly for low values of $k$ but outperform CLIP as the value of $k$ increases. The result indicates that the overhead embeddings from the Sat2Cap model are more diverse, allowing them to learn a variety of fine-grained concepts. The Sat2Cap embeddings can be well separated into a large number of clusters (concepts) and do not suffer the same issue of collapse as CLIP, making them suitable for the task of fine-grained mapping. 
\begin{figure*}[!t]
\begin{center}
\includegraphics[width=0.95\linewidth]{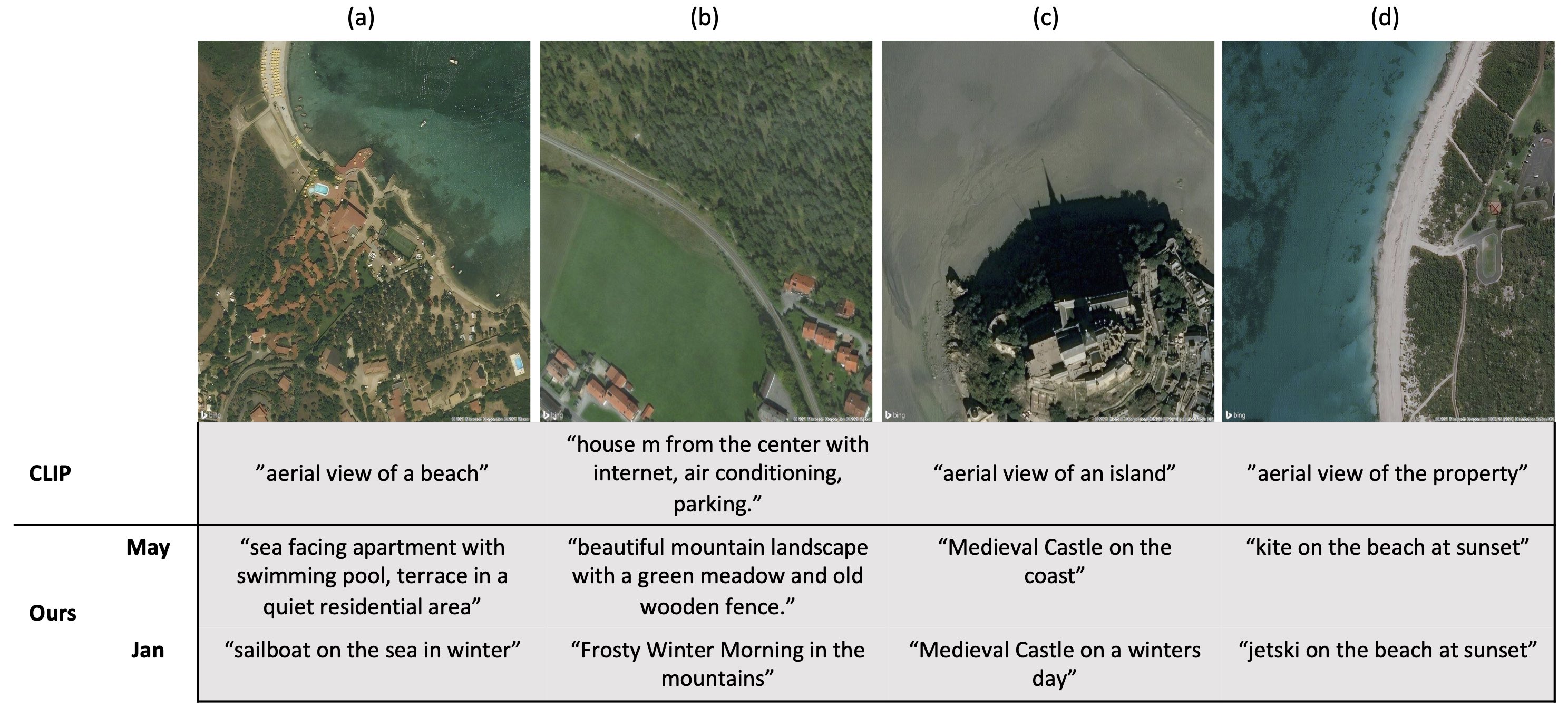}
\end{center}
   \caption{Captions generated by the CLIPCAP model \cite{mokady2021clipcap} using CLIP embeddings vs.\ Dynamic Sat2Cap embeddings. (Row-1) shows the results from CLIP embeddings, which produce many generic descriptions. (Row-2 and 3) show the results from our Sat2Cap embeddings for the month of May and January, respectively. The captions generated using Sat2Cap embeddings are more fine-grained and dynamic. While (d) does not add any winter properties for the January query, this behavior is expected as the image is over Australia, where January falls in the middle of summer.}    
\label{fig:img1}
\end{figure*}

\begin{table*}
\begin{center}
\begin{tabular}{c|ccc|cc|c}
\hline
\multirow{2}{*}{} & Meta   & \multirow{2}{*}{Dropout} & Meta   & \multicolumn{2}{c|}{Cosine Similarity} & \multirow{2}{*}{BERT Score} \\ \cline{5-6}
     & Training &        & Inference & MPNET\_Base\_v2 & E5\_Small &        \\ \hline
CLIP & -        & -      & -         & 0.1560 & 0.7519 & 0.7135 \\
\hline
ours              & \xmark & \xmark                   & \xmark & \textbf{0.3334}    & \textbf{0.7969}   & \textbf{0.7692}             \\
     & \cmark   & \xmark & \xmark    & 0.2727 & 0.7806 & 0.7476 \\
     & \cmark   & \xmark & \cmark    & 0.2888 & 0.7857 & 0.7572 \\
     & \cmark   & \cmark & \xmark    & 0.2755 & 0.7831 & 0.7533 \\
     & \cmark   & \cmark & \cmark    & 0.2755 & 0.7833 & 0.7537
\end{tabular}
\end{center}
\caption{\textbf{Caption generation alignment:} The table shows the alignment between captions generated from co-located overhead and ground-level images. Captions generated using Sat2Cap have better alignment with the descriptions of co-located ground-level scenery.}
\label{table:text_metrics}
\end{table*}
\subsection{Zero-Shot Map of Fine-grained Concepts}
\label{sec:zero_shot}
We use Sat2Cap embeddings to create zero-shot maps using fine-grained prompts. To do this, we choose a region and download high-resolution satellite images $(0.6m/px)$ over the region. We precompute the Sat2Cap embeddings for all the images.
At inference time, for a given text query, we predict the CLIP embedding and compute its similarity with the overhead image embeddings. The process of computing the similarities only takes about 4 seconds. 

Figure~\ref{fig:fg_maps} shows the zero-shot map of Amsterdam for two prompts: ``Heavy trucks transporting goods", and ``People walking in the streets of downtown." Column 1 shows the maps generated using CLIP embeddings.
For the first prompt, the Sat2Cap activations are localized to the highway network of the city, whereas CLIP has many false positives over general built-up areas. This shows that Sat2Cap can pick the subtle nuances of fine-grained text that CLIP fails to do. Similarly, for the second prompt, CLIP shows high activations throughout the city, while Sat2Cap's activations are more localized to the downtown area.  

We also create country-level zero-shot maps for two countries: the Netherlands and England. We query the images with 3 prompts, each related to a distinct land cover class. Figure~\ref{fig:generic_maps} shows the comparison of our zero-shot maps with the land cover maps obtained from ESRI. We see that the zero-shot maps highly correlate with the ESRI land cover maps of the respective countries.

\subsection{Fine-grained and Dynamic Caption Generation}
To generate captions from our embeddings, we use the CLIPCap~\cite{mokady2021clipcap} model. CLIPCap allows us to generate captions by learning a mapping from the CLIP space to the text space. Figure~\ref{fig:img1} shows qualitative examples of captions generated by passing CLIP embedings vs Sat2Cap embeddings as input for a given overhead image. We observe that when using CLIP embeddings of the overhead images as input, the captions generated by CLIPCap mostly describe generic concepts of a location like a beach, island, property, etc. In contrast, our Sat2Cap embeddings produce more fine-grained and aesthetically pleasing captions. Furthermore, we use the dynamic encoder to generate captions in two different months, for the same location. Figure~\ref{fig:img1} shows that Sat2Cap accurately models the seasonal variations and aligns the captions towards respective temporal inputs such as capturing winter concepts for January. However, in Figure (d), we see that the model does not add any winter-specific information for the January input. This is expected behavior since the image is from Australia, where the month of January falls in the middle of summer. The example further demonstrates that Sat2Cap learns a joint model of time and location.
Fine-grained captions for a location should be well-aligned with the captions describing the ground-level scene. We quantitatively evaluate the quality of the captions by looking at their alignment with the text descriptions for respective ground-level images. First, we use the CLIPCap model to generate captions for ground-level images and use that as our ground truth. We then use the CLIP embeddings and Sat2Cap embeddings of the overhead images to generate the respective captions for that location. Table~\ref{table:text_metrics} shows the similarity metrics between the ground-truth text and generated text. To compute the cosine similarity, we use two different sentence transformers from HuggingFace, \textit{MPNET\_Base\_v2}~\cite{song2020mpnet} and \textit{E5\_Small}~\cite{wang2022text}. The captions generated using Sat2Cap embeddings demonstrate significantly better alignment with the ground-image captions than their CLIP counterparts. The results show that Sat2Cap captions better describe the ground-level characteristics of a given location. Table~\ref{table:text_metrics} also shows that the model trained without metadata has the highest cosine similarity and BERTScore with ground-level descriptions. We suspect this happens because of the uncertainty introduced by the use of a pretrained caption generator, i.e., small noise in the metadata introduces big deviations in the generated text. 


%% file: sec/5_conclusion.tex
\section{Conclusion}
\label{sec:conclusion}

We presented a weakly supervised framework for learning a semantically rich embedding space between geolocation and fine-grained text. For this task, we introduced a new large-scale cross-view dataset with 6.1M samples. Our approach does not depend on text supervision, and learns textual representations of a location using only ground-level images and the CLIP embedding space. In addition to higher retrieval performance, we demonstrated that our framework can efficiently generate high-quality zero-shot maps from fine-grained text prompts. This ability to create maps from fine-grained text prompts offer greater flexibility when compared to traditional methods. Finally, we demonstrated that Sat2Cap embeddings can be used to generate dynamic captions that align with the ground-level scene. 


%% file: sec/X_suppl.tex
\clearpage
\setcounter{page}{1}
\maketitlesupplementary

\section{Implementations Details}
We use a ViT-32B as the CLIP image encoder. This image encoder is kept frozen throughout the training. We use a ViT-32B architecture as the backbone for our Sat2Cap model. The Sat2Cap backbone is initialized using CLIP weights. Following~\cite{radford2021learning} we use an AdamW optimizer~\cite{loshchilov2017decoupled} with a learning rate of $1e-05$ with $\beta_1=0.9$ and $\beta_2=0.98$. We use a learnable temperature parameter initialized at $\tau=0.07$. We use Cosine Annealing with Warm Restarts~\cite{loshchilov2016sgdr} as the learning rate scheduler. We augment the overhead images using RandomResizedCrop and RandAugment~\cite{cubuk2020randaugment}. The overhead images are normalized using the mean and standard deviation of the training set. The training was carried out using Nvidia A100 40GB GPU. Since a larger number of negative samples is beneficial for contrastive learning, we simulate a large batch size using a memory bank approach. We initialize a queue of size 9600 and fill it with precomputed ground-level image CLIP embeddings which are used as negative samples for computing the loss.  

\section{Text to Overhead Image Retrieval}
\begin{figure*}[t]
\begin{center}
\includegraphics[width=0.9\linewidth, height=1.2\linewidth, scale=0.3]{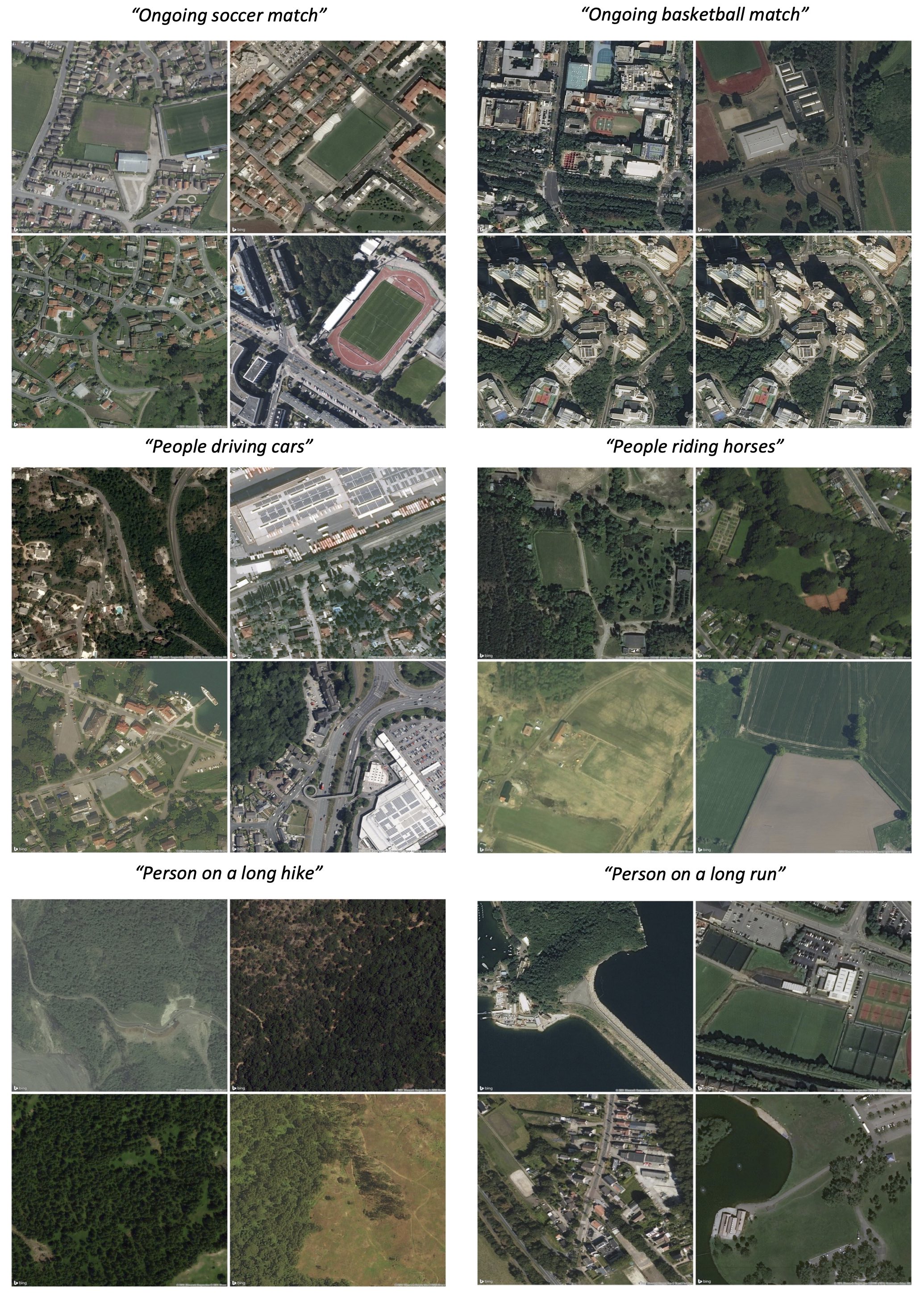}
\end{center}
   \caption{\textbf{Top-4 text-to-overhead retrieval:} We retrieve the top-4 closest overhead image from a given text prompt. Our results show that Sat2Cap embeddings can accurately relate geolocations with fine-grained textual prompts.}  
\label{fig:text2ov}
\end{figure*}

Our framework uses ground-level images as pseudo-labels to learn the textual concepts of geolocation. Although Sat2Cap does not require any text labels during training, it effectively learns an embedding space where geolocation and their fine-grained descriptions are well aligned. To show this, we randomly selected 1000 overhead images from our training set, and compute their Sat2Cap embeddings. For a given text query, we generate the CLIP\cite{radford2021learning} text embedding and compute its similarity with all images in the test set. Figure~\ref{fig:text2ov} shows examples of 4 closest overhead images retrieved for a given query.

We experiment with small perturbations of prompts to analyze how our retrieval results change with minute variations of query. We see in Figure~\ref{fig:text2ov}, the prompt ``people driving cars" retrieves city or residential areas. However, replacing the phrase ``driving cars" with ``riding horses" retrieves locations with farmland. Similarly, the prompt ``person on a long hike" exclusively retrieves mountainous regions, while the prompt ``person on a long run" retrieves images that looks like trails nearby residential areas. Hence, Sat2Cap embeddings demonstrate a good understanding of fine-grained variations of textual concepts.

\section{Learning Dynamic Concepts}
\begin{figure*}[t]
\begin{center}
\includegraphics[width=0.95\linewidth]{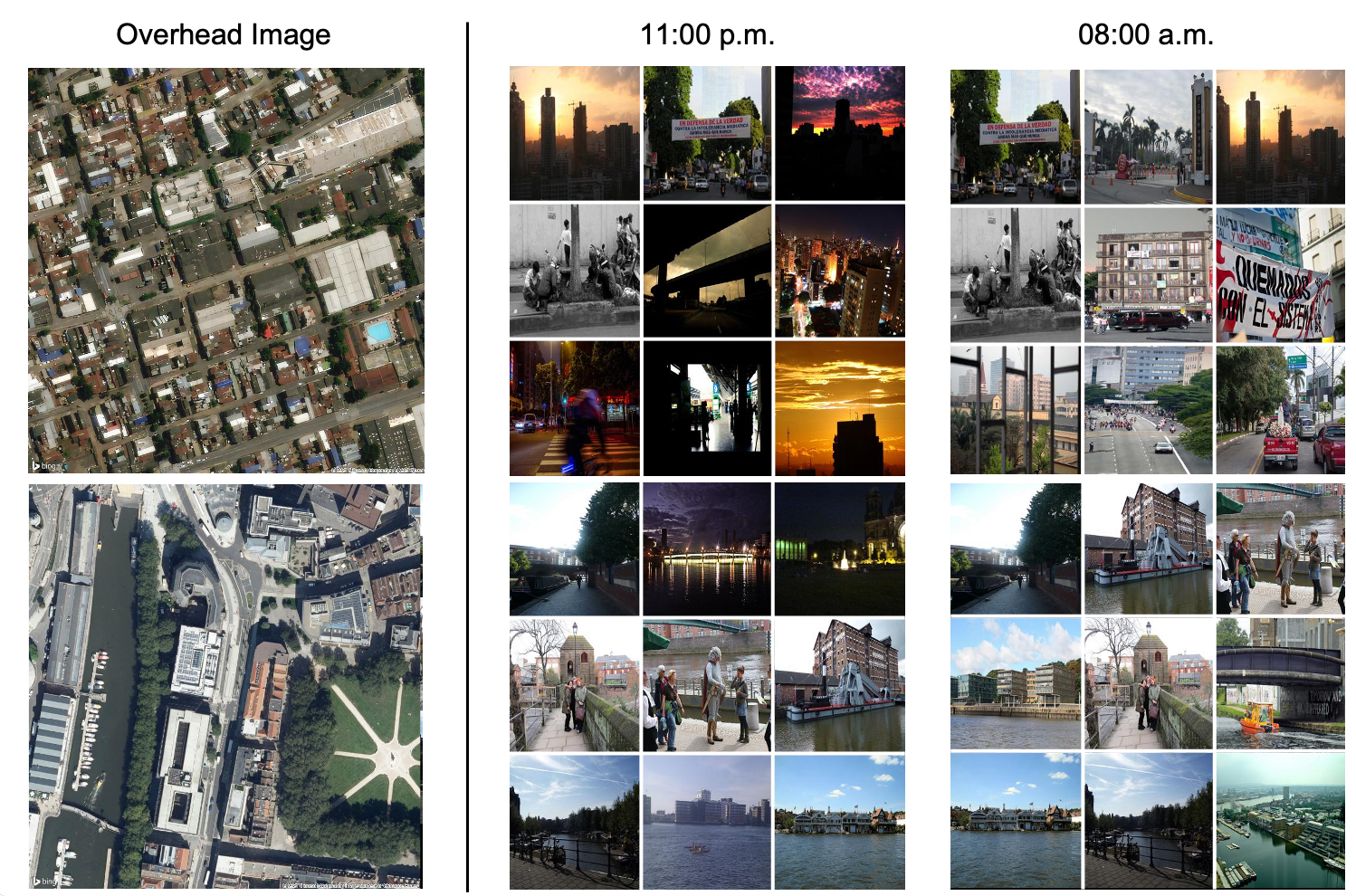}
\end{center}
   \caption{\textbf{Top-9 overhead-to-ground image retrieval with temporal manipulation:} We show the 9 closest ground-level images for a query overhead image at two different time settings (11:00 p.m. and 08:00 a.m.).}  
\label{fig:dynamic_retrieval}
\end{figure*}

\begin{figure*}[t]
\begin{center}
\includegraphics[width=0.9\linewidth, scale=0.3]{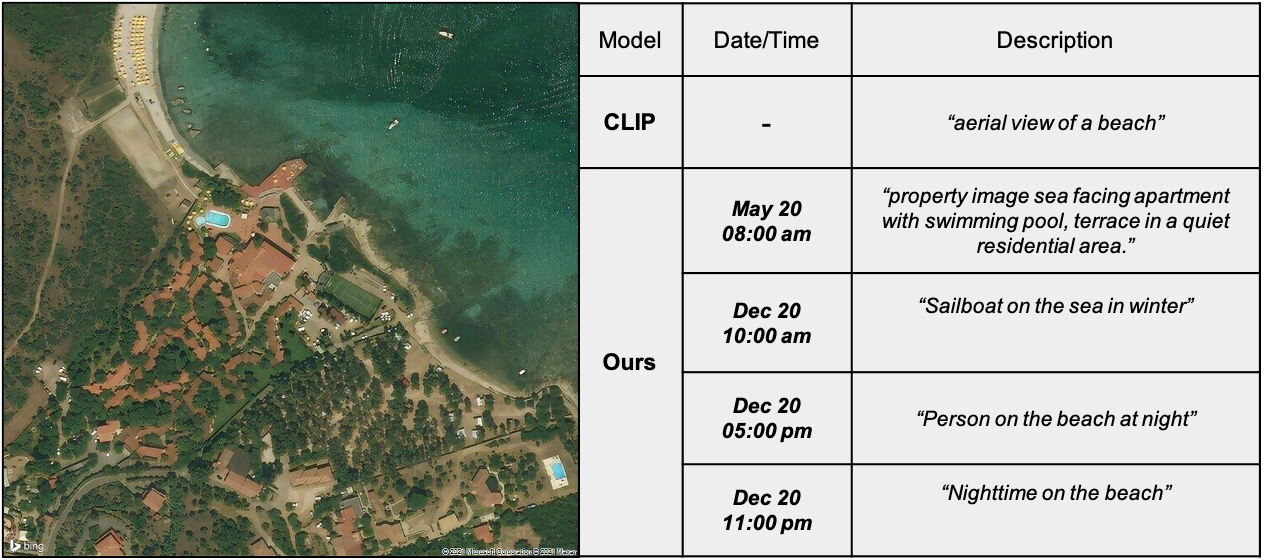}
\end{center}
   \caption{\textbf{Dynamic Caption Generation:} Our Sat2Cap embeddings dynamically adapt to temporal manipulations, facilitating dynamic caption generation.}  
\label{fig:dynamic_caption}
\end{figure*}

\begin{figure*}
\begin{center}
\includegraphics[width=0.9\linewidth, scale=0.3]{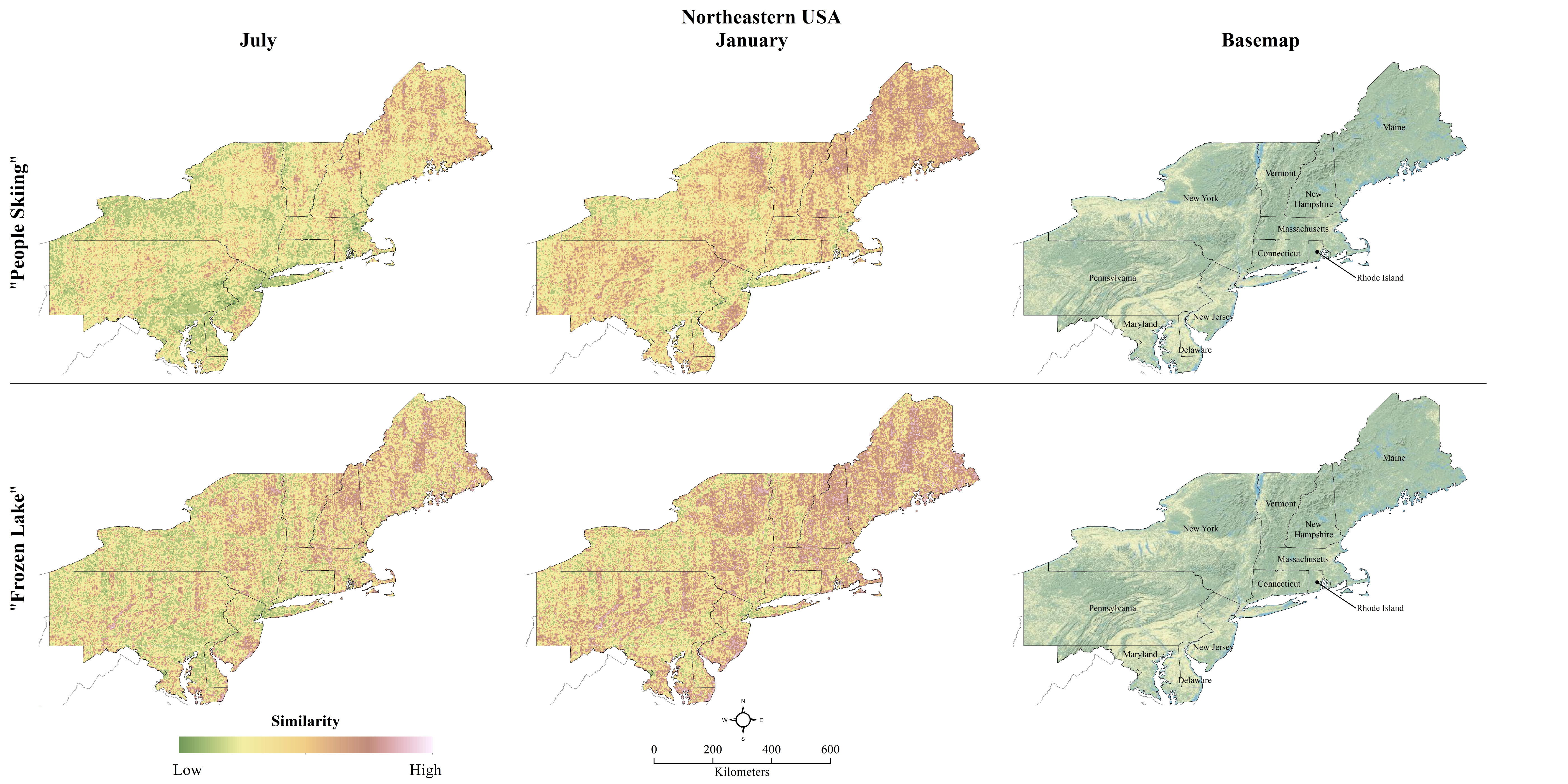}
\end{center}
   \caption{\textbf{Dynamic Maps:} We show zero-shot maps of the northeast US at two different temporal settings (July and January)}  
\label{fig:dynamic_map}
\end{figure*}

The dynamic encoder allows our model to learn temporally varying concepts over a location. Here we show more qualitative results showcasing the dynamic properties of our model. Figure~\ref{fig:dynamic_retrieval} shows the retrieval results at two different time settings (11:00 p.m. and 08:00 a.m.). Figure~\ref{fig:dynamic_caption} shows the captions generated for the same location over long (order of months) and short (order of hours) term variations. Finally, Figure~\ref{fig:dynamic_map} shows how our model can generate maps that adapt to temporal variations for a given prompt.

\section{Country Level Map of US}
\begin{figure*}
\begin{center}
\includegraphics[width=0.9\linewidth,scale=0.3]{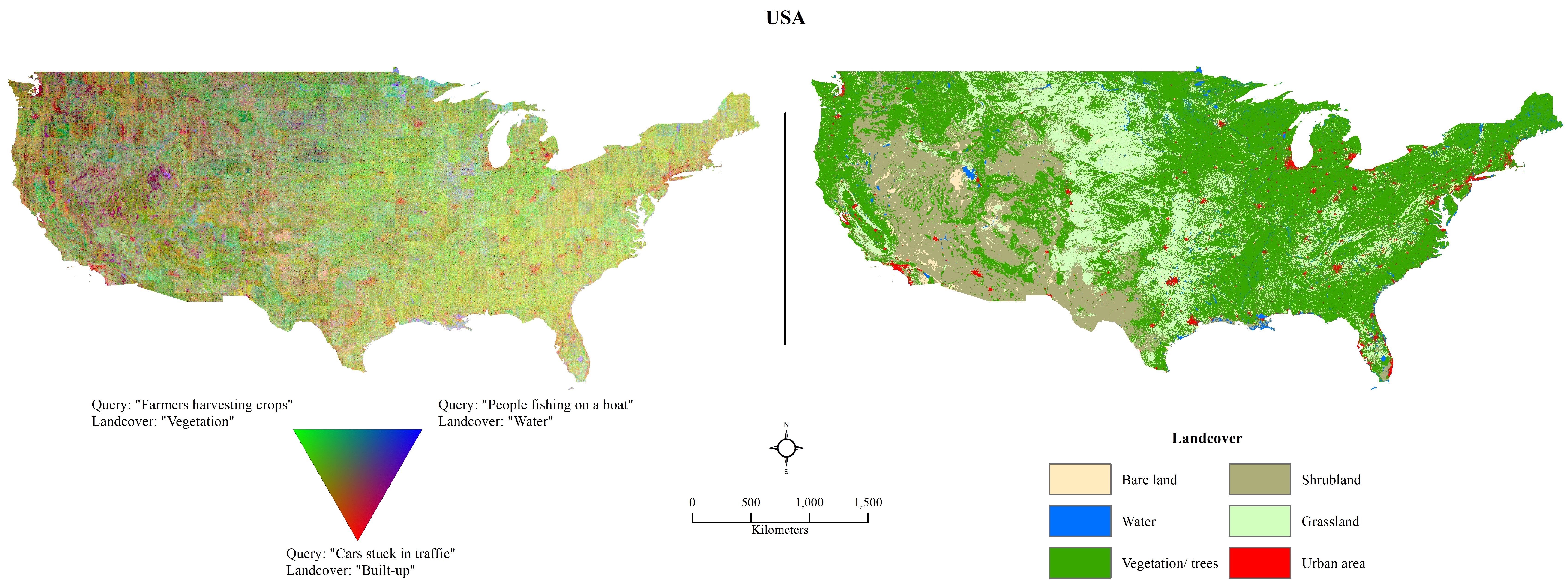}
\end{center}
   \caption{Zero-shot map of the USA}  
\label{fig:usa_map}
\end{figure*}

We also create a zero-shot map of the US. However, due to the massively large area, we had to downsample our overhead image acquisition by 10x. Although we are predicting at a much coarser resolution, we still achieve a reasonable zero-shot map as seen in Figure~\ref{fig:usa_map}.

\section{Geolocalizing Textual Queries}
\begin{figure*}[t]
\begin{center}
\includegraphics[width=0.95\linewidth]{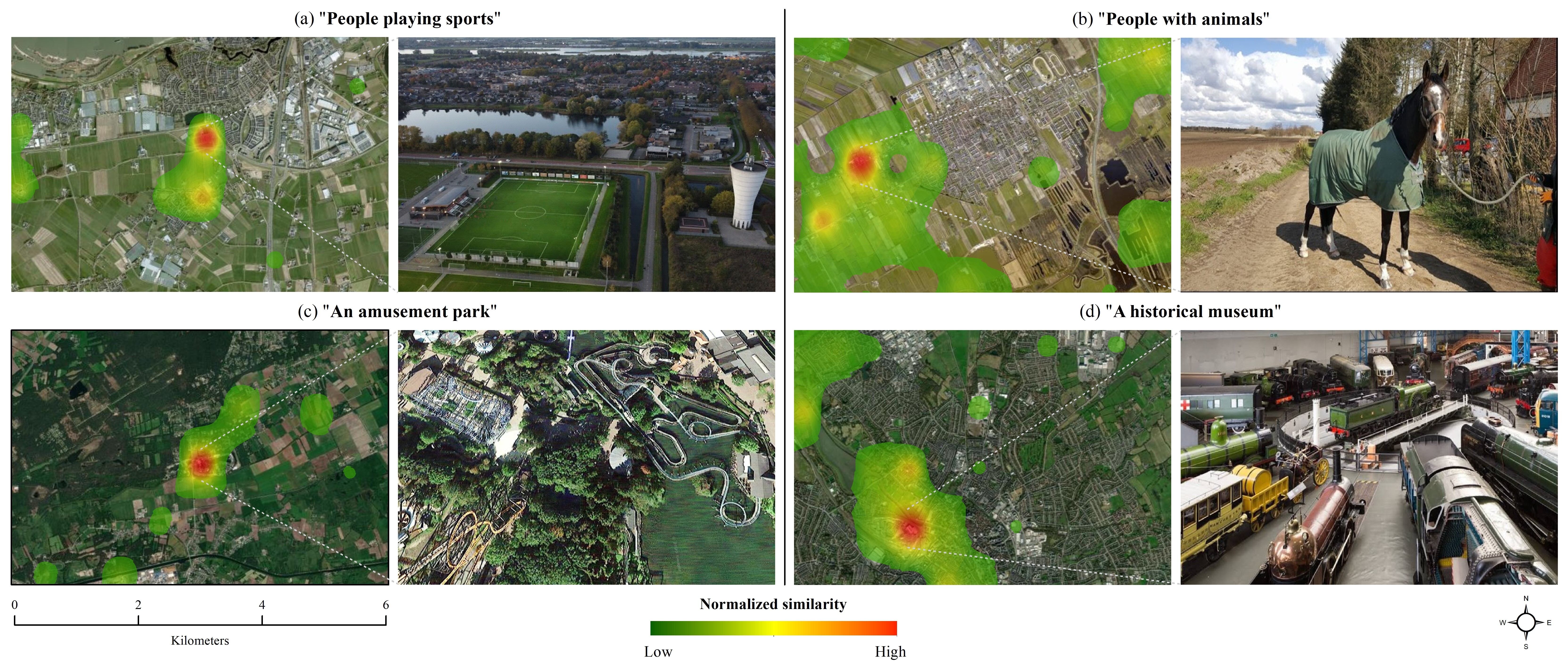}
\end{center}
   \caption{\textbf{Localizing textual queries at finer resolution}: For each prompt, the image on left shows the big region which is used for inference. The image on the right shows an image of the ground-level scene at the point with the highest activation, which was taken by entering the location in Google Maps}  
\label{fig:localize}
\end{figure*}

Our model can be used to localize textual queries at a finer resolution. For this experiment, we draw a $24 km^2$ bounding box over a region. We compute the Sat2Cap similarity for all the overhead images in that box with a given text query. We, then, normalize the similarities between $0$ and $1$ and clip the values below $0.5$. 
Figure~\ref{fig:localize} shows the results of this experiment. The red spot indicates the location with the highest activations for the given query. For each query, the left figure shows the total area of inference, and the right figure shows a fine-grained image at the location with the highest activation, obtained from Google. We see that our model makes reasonable localization for the given queries. For example: in (a) our model activates over a soccer stadium. Figure (b) shows that when we compose the concept of people with animals, our model shows very high activation in farm-like areas which is where these two concepts would most likely co-occur. 
These results show that our model can reasonably localize the most plausible point within a given area, where one might observe a given query. This property can be beneficial in solving visual search problems in the geospatial domain.

\section{Dataset}
\begin{figure*}
\begin{center}
\includegraphics[width=0.9\linewidth, scale=0.3]{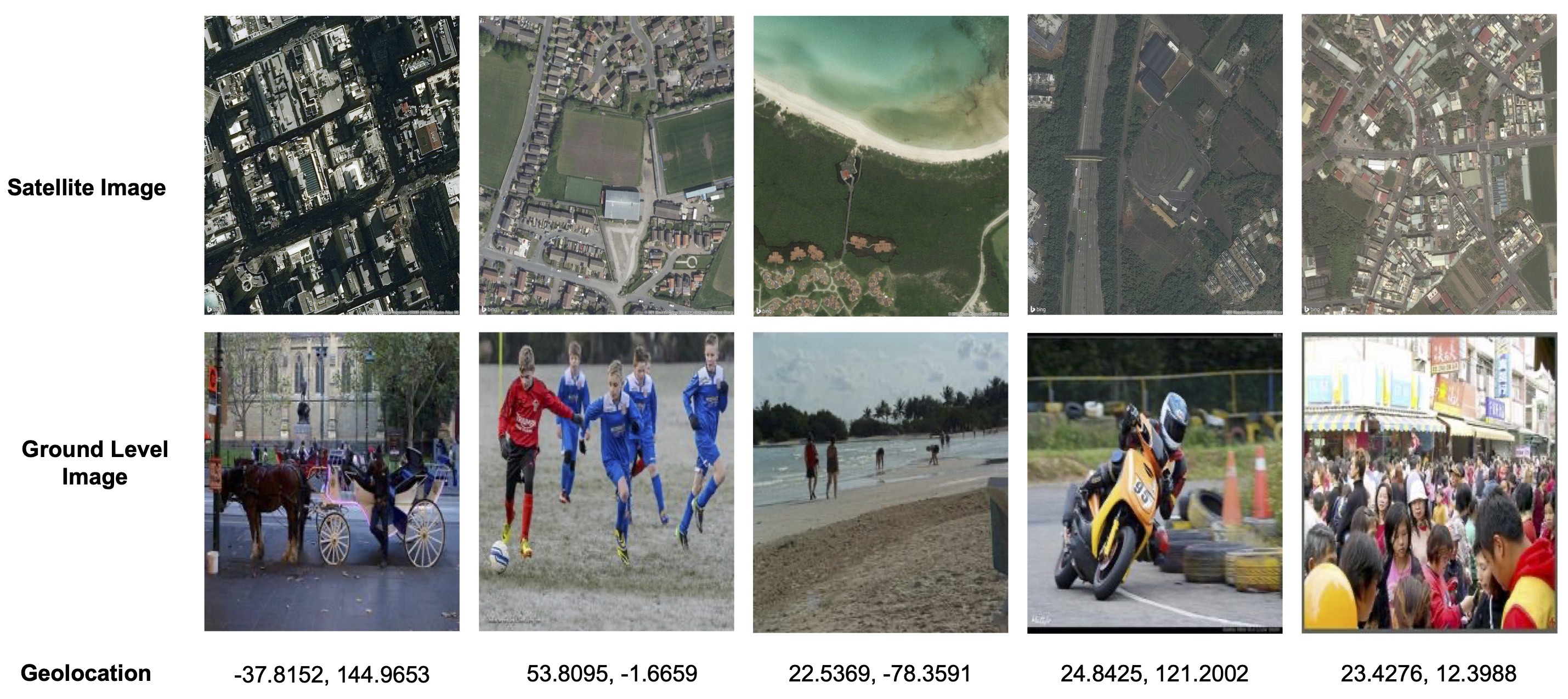}
\end{center}
   \caption{Examples of co-located overhead and ground images in our dataset. The ground-level images describe more detailed concepts of the given locations than their overhead counterparts.}  
\label{fig:dataset}
\end{figure*}

We introduced a cross-view dataset with overhead images and co-located ground-level images taken from the YFCC100M~\cite{thomee2016yfcc100m} dataset. Figure~\ref{fig:dataset} shows a few samples from our dataset. The ground-level images provide us with detailed fine-grained concepts of a location that cannot be directly inferred when looking at the overhead imagery.


\begin{figure*}[!t]
\begin{center}
\includegraphics[width=0.95\linewidth, height=1.1\linewidth,scale=0.3]{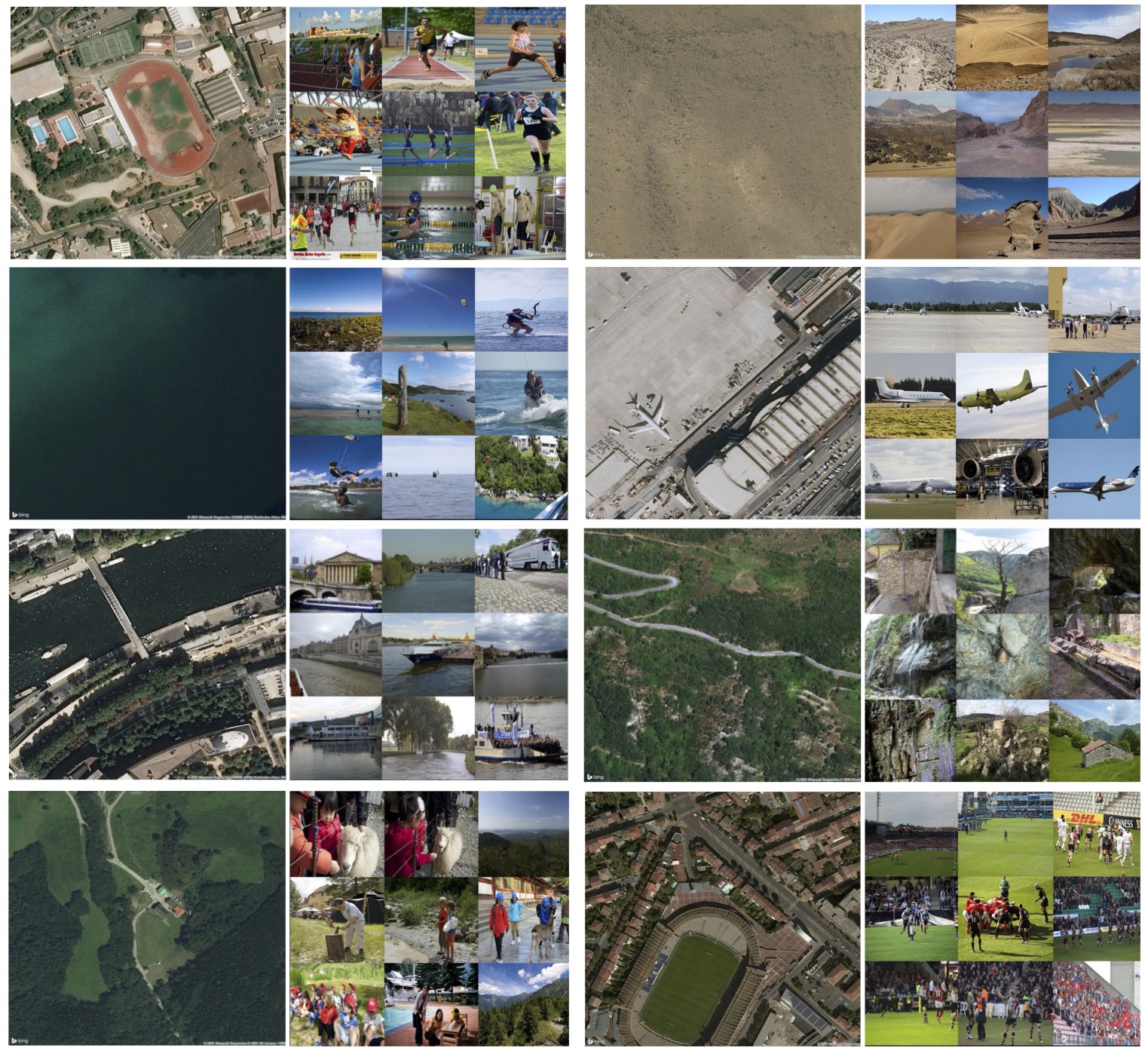}
\end{center}
   \caption{\textbf{Top-9 overhead-to-ground retrieval}: Our model can infer fine-grained concepts of ground-level scenes through overhead imagery. Sat2Cap accurately retrieves probable concepts for a given geolocation using an overhead image.}  
\label{fig:ov2gr}
\end{figure*}